\def\year{2020}\relax
\documentclass[letterpaper]{article}
\usepackage{aaai20}
\usepackage{times}
\usepackage{helvet}
\usepackage{courier}
\usepackage[hyphens]{url}
\usepackage{graphicx}
\usepackage{comment}

\graphicspath{{figures/}}

\usepackage{amssymb}

\title{CG-GAN: An Interactive Evolutionary GAN-based Approach \\for Facial Composite Generation}

\author{Nicola Zaltron \\ IT University of Copenhagen \\ Copenhagen, Denmark \\ nicolazaltra@gmail.com
\And Luisa Zurlo \\ IT University of Copenhagen \\ Copenhagen, Denmark \\ zurlo.luisa@gmail.com
\And Sebastian Risi \\ IT University of Copenhagen \\ Copenhagen, Denmark \\ sebr@itu.dk}

\begin{document}
\maketitle

\begin{abstract}
Facial composites are graphical representations of an eyewitness's memory of a face. Many digital systems are available for the creation of such composites but are either unable to reproduce features unless previously designed or do not allow holistic changes to the image. In this paper, we improve the efficiency of  composite creation by removing the reliance on expert knowledge and letting the system learn to represent faces from examples.
The novel approach, \emph{Composite Generating GAN} (CG-GAN), applies generative and evolutionary computation to allow casual users to easily create facial composites. Specifically, CG-GAN utilizes the generator network of a pg-GAN to create high-resolution human faces. Users are provided with several functions to interactively breed and edit faces.  CG-GAN offers a novel way of generating and handling static and animated photo-realistic facial composites, with the possibility of combining multiple representations of the same perpetrator, generated by different eyewitnesses.

\end{abstract}

\noindent 

\begin{figure}[tbh!]
    \centering
    \includegraphics[width=0.9\linewidth]{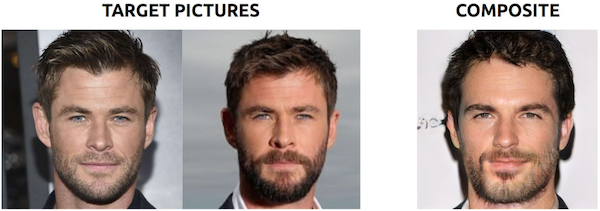}
    \caption{Example composite built using CG-GAN.}
    \label{fig:intro_example}
\end{figure}

\section{Introduction}

Facial composites are portrait sketches of unknown individuals used in criminal investigation to identify a person. Two cognitive abilities are applied in the process: recall and recognition. Recall is the recollection of information for the composite construction. Recognition is the ability to recognize someone seen before, used during line-ups \cite{facialCompositeSystemReview,pcs}. Compared to recall, recognition is an easier and stronger task, 
therefore most courts of law consider lineup identification as good evidence \cite{pcs}. 

Originally composites were drawn by forensic artists in consultation with victims and eyewitnesses, relying on detailed descriptions of their memories.
More recent approaches involved mechanical and digital systems to improve the process and the resulting recognition rate. 

There exist two major modern creation techniques. Simple systems merge single features (drawn or picked from a feature set) into a portrait. Others let the witnesses concentrate on the entire face, by selecting individuals as in a line-up \cite{IE_composites_article}.
This \textit{holistic approach} is the result of psychological research suggesting that describing individual features causes lower identification rates \cite{IE_composites_article}. 

CG-GAN is designed to make the process artist-free to avoid communication issues and potentially lower retention time, and is empowered by the usage of recent machine learning techniques. In more detail, the approach is based on the \emph{Latent Variable Evolution} (LVE) approach \cite{bontrager2017deepmasterprint}, in which a Generative and Adversarial Network (GAN) is trained on a specific dataset and the space of images encoded by the GAN is then searched through an evolutionary computation approach. LVE has been applied to diverse domains such as the  generation of fingerprints \cite{bontrager2017deepmasterprint}, image generation \cite{deepInteractiveEvolution} and even the creation of Mario levels \cite{volz2018evolving}.

\citeauthor{deepInteractiveEvolution}~\shortcite{deepInteractiveEvolution} showed that 
LVE can  enable users to guide the search  through a process known as interactive evolution  \cite{takagi2001interactive}. However, evolving towards a specific target image, which is a prerequisite for facial composite generation,  has shown challenging \cite{deepInteractiveEvolution}. This work goes beyond the previous state-of-the-art by extending LVE with the ability to freeze certain features discovered during the search, enabling a more controlled user-recreation of target images. 
The system exploits advantages of both constructive and holistic techniques.
Figure~\ref{fig:intro_example} shows an example of a composite session that has been carried out with the usage of CG-GAN.

\section{Related work}

\subsection{Facial Composite Generation} 
Police agencies are already using software for the creation of facial composites, such as Faces 4.0, which allows adding and editing features \cite{faces}. However, this kind of software is not capable of reproducing facial features that have not been previously designed, is not holistic and requires the user to explore a vast feature database to construct a face.  Other programs such as EvoFit \cite{evofit_paper} and EFIT-V \cite{efitV} are based on an interactive evolutionary algorithm. These systems can produce high-quality composites but require complex mathematical functions to create the face. Every aspect that is not planned a priori can therefore not be produced. 

Interactive evolutionary computation (IEC) is a particular form of AI-assisted creation, in which a human functions as the fitness function of an evolutionary algorithm. IEC is often used in subjective domains which 
are hard to define mathematically, such as in
evolving two-dimensional images \cite{secretan2011picbreeder}, three-dimensional forms \cite{clune2011evolving}, musical
compositions \cite{hoover2012generating}, or agent behaviors \cite{de2017interactive}.   
In an iterative fashion, the system presents a number of artifacts for the experimenter to evaluate; the user, in turn,  responds by indicating which of these artifact(s) is preferred. The next generation of artifacts is produced through mutations and/or crossover to the underlying representation, from which the user selects their favorite artifact(s), again forming the basis for the next generation.  While the conceptual idea of IEC is intriguing, there are several challenges hampering the practical usefulness of the technique. Often a large number of evaluations are needed to find the desired artifact,  an issue known as user fatigue. Additionally, it can be difficult to find a particular artifact a user has in mind. Part of the problem can be traced back to the underlying representation, i.e.\ the  employed genotype-to-phenotype mapping, which might thread the wrong balance between generality and domain specificity.  

\subsection{Generative models}

Generative models are a branch of machine learning techniques, in  which the objective is to generate content \cite{lecun2015deep}. They include diverse image-related application areas such as  completion, correction, production of variations, denoising, up-scaling, etc. In the work presented here, we employ Generative Adversarial Networks (GANs) \cite{originalGanPaper}, which are a class of ML algorithms that are trained in an unsupervised way. GANs make use of two neural networks (NN) to simultaneously train two models: a generator $G$ estimates the data distribution while a discriminative model $D$ estimates if samples are from the training set or synthetic. During training, the goal of $G$ is to maximize the probability of $D$ misclassifying samples,  like in a minimax two-player game. Eventually, $G$ recovers the data distribution and $D$ becomes unable to distinguish generated content.

Many extensions have been proposed, such as Deep Convolutional Generative Adversarial Networks (DCGANs)~\cite{radford2015unsupervised}, a class of Convolutional Neural Networks (CNNs); Auto-Encoder Generative Adversarial Networks (AE-GANs)~\cite{makhzani2015adversarial}; and Plug and Play Generative Networks (PPGNs)~\cite{nguyen2016plug}. 
This paper employs a  recent extension to GANs called Progressive Growing of GANs (pg-GAN) \cite{pgGans}, which is a training methodology for GANs that allows generator and discriminator to grow progressively: Starting  with low (4 $\times$ 4) spatial resolution, new layers are added  as the training proceeds to model increasingly fine details \cite{pgGans}. This speeds up and stabilizes training, allowing the creation of high-quality images.

In this paper, we employ a  pg-GAN\footnote{\url{https://github.com/tkarras/progressive_growing_of_gans}} pre-trained on the CelebFaces Attributes Dataset \cite{celeba}, which contains 200,000 images of celebrity faces, annotated with 40 binary attributes. The dataset (or its HQ version; 
\citeauthor{pgGans}, \citeyear{pgGans}) can be used for learning both the creation of facial images \cite{pgGans,style_gan_article,tlGan} and assigning attributes to face images \cite{tlGan}.

\begin{figure}[t]
    \centering
    \includegraphics[width=0.6\linewidth]{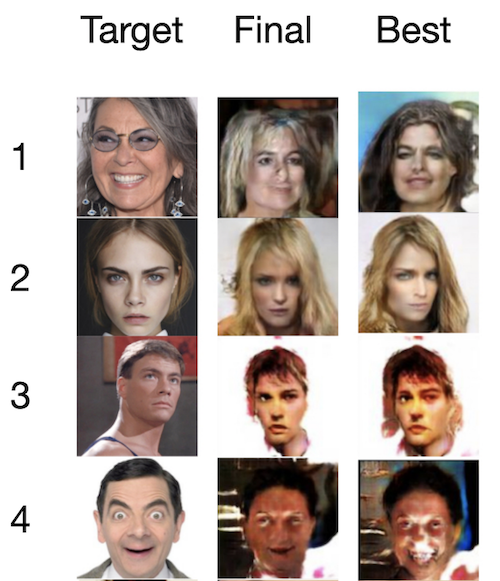}
    \caption{Latent variable evolution results from \cite{deepInteractiveEvolution}. }
    \label{fig:previous}
\end{figure}
\subsection{Latent Variable Evolution}
\begin{figure}
    \centering
    \includegraphics[width=0.965\linewidth]{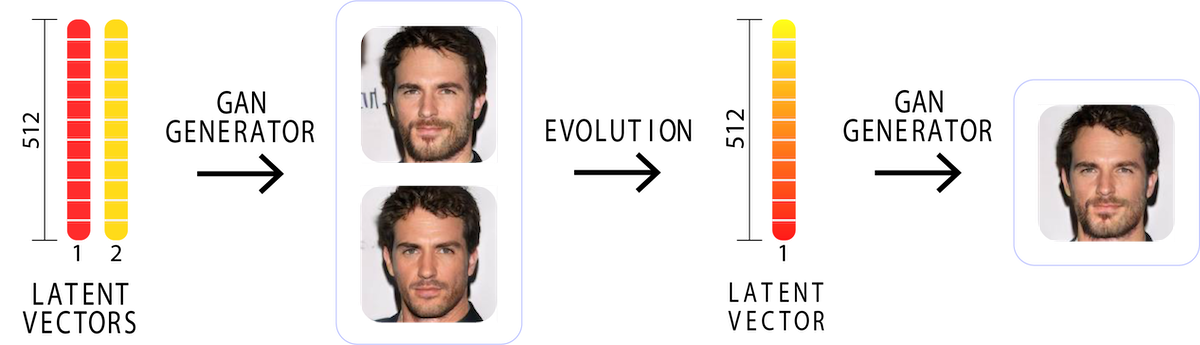}
    \caption[An example of how the evolution works.]{Latent Variable Evolution. Starting from a pre-trained GAN (left), the latent vector space (right) can be searched to create images with certain properties.}
    \label{fig:workflow1}
\end{figure}

The approach in this paper is based on \emph{latent variable evolution} (LVE); the basic idea is shown in Figure~\ref{fig:workflow1}. First a GAN is trained in an unsupervised way to generate specific target content. In the second phase, the space of content can be searched by applying evolutionary techniques to the latent vector fed into the GAN's generator. The first LVE approach was introduced by \citeauthor{bontrager2017deepmasterprint}~\shortcite{bontrager2017deepmasterprint}. In their work, the authors train a GAN on a set of real fingerprint images and then apply evolutionary search to find a latent vector that matches with as many subjects in the dataset as possible.

In another paper \citeauthor{deepInteractiveEvolution}~\shortcite{deepInteractiveEvolution} present an interactive LVE system, in which users can evolve the latent vectors for a GAN trained on different classes of objects such as faces or shoes (an approach also employed by the popular Ganbreeder app \cite{ganbreeder}).  Because the GAN is trained on a specific target domain, it becomes a compact and robust genotype-to-phenotype mapping (i.e.\ most produced phenotypes do resemble valid domain artifacts) and users were able to guide evolution towards images that sometimes resembled given target images. Such target-based evolution has been shown to be challenging with other generative  representations  \cite{on_the_deleterious_effects}. However, the approach introduced by \citeauthor{deepInteractiveEvolution}~\shortcite{deepInteractiveEvolution} does not allow  to freeze discovered facial features,  which limits the amount of control the user has during the search (Figure~\ref{fig:previous}). For example, once a facial feature such as the beard looks just right and the user would only like to tweak the eyes, subsequent mutations to the latent vector will likely change both of these features. 

\section{Approach: CG-GAN}
\begin{figure}[tbh!]
    \centering
    \includegraphics[width=1\linewidth]{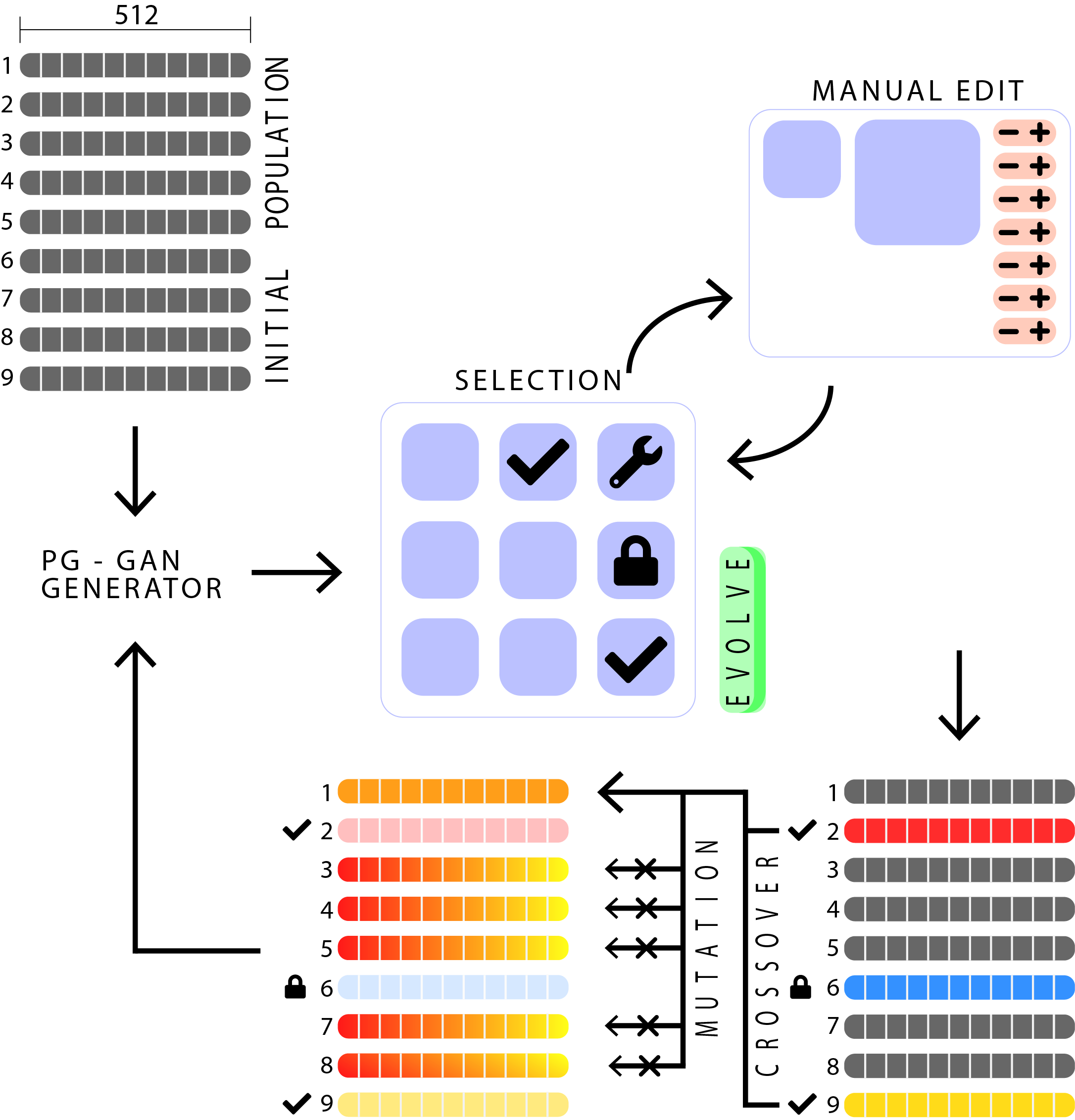}
    \caption{CG-GAN overview. The approach starts by presenting the user with a selection of varied images by initially inputting random latent vectors in a pg-GAN generator. Next the user can evolve faces interactively, manually edit, or lock features. The process is repeated for a number of generation, until the desired composite is created.}
    \label{fig:intro_overview}
\end{figure}

CG-GAN combines GANs with interactive evolution and additional features for fine-grained control  (Figure~\ref{fig:intro_overview}). Through IEC  users are able to explore solutions and control the input (i.e.\ the latent vector) of a pre-trained GAN's generator. Similarly to previous work by \citeauthor{deepInteractiveEvolution}~\shortcite{deepInteractiveEvolution}, this approach limits user fatigue by restricting the search space to the learned  genotype-to-phenotype of the generator. While the approach by \citeauthor{deepInteractiveEvolution}~\shortcite{deepInteractiveEvolution} implements a simple interface to visualize and select individuals, with a slider acting on the standard deviation of the mutation's noise, the approach introduced in this paper gives users fine-grained control. This advance enables the evolution of accurate composite sketches and is realized through an intuitive user interface and two essential approach additions: feature extraction and smart locks. 

\textbf{Feature extraction.} One challenge with GANs is  the control of the output: although the synthetic content can have remarkable quality, it  depends on the random noise that is fed into the generator. To gain precise control of individual composite features, our approach includes a method called Transparent Latent Space GANs (TL-GAN) \cite{tlGan}, which consists in training a \textit{classifier} CNN to correlate the changes in the input space to the changes in the output. Here facial features are measurable and representable characteristics of human faces. Facial features represent the major mean for communicating the traits recalled by eyewitnesses and for comparing and describing differences.
Moreover, such features are essential for manually editing faces and can also enable a more controlled evolutionary process. Indeed, features can be interpreted as axes along which any face can be modified, both manually or automatically, to edit specific traits of a given face. The approach, based on TL-GAN, consists in discovering feature axes in a trained generator's latent space so that a vector can move along an axes to morph an image's feature. This is achieved by finding correlations between noise vectors and image features, through supervised learning \cite{tlGan}. For CG-GAN, we re-created our own models and feature axes. 

\textbf{Feature locks and smart locks.} 
An issue with editing feature axes is that one change can potentially modify other correlated features (e.g.\ decreasing beard makes the face more female-like).
As in TL-GAN, we disentangle features (i.e.\ axis in  multidimensional space) by orthogonalizing axes, using their projections over different directions \cite{tlGan}. In other words, components that interfere with that axis are subtracted, which creates a new axis that is somewhat similar to the non-orthogonalized one. CG-GAN allows users to lock features: when a feature gets locked all the other ones are re-calculated by subtracting their projection over locked axes, thus avoiding their interference with any of the currently locked ones. CG-GAN also introduces \textit{smart locks} to assist the user by locking sets of features based on their correlation. In more detail, the cosine similarity is computed for each pair of axes and when a feature is smart-locked both that feature and all the ones \textit{strongly correlated} to it (here defined as exceeding a cosine similarity threshold of $\pm0.5$)  are locked/unlocked. This addition allows  users to modify traits without overwriting others that are likely to change. As an example, by \textit{smart-locking} the gender attribute,  also beard, moustache, hair length, makeup are locked, among others.

\subsection{CG-GAN User Interface}

\begin{figure}[htb!]
    \centering
    \includegraphics[width=1\linewidth]{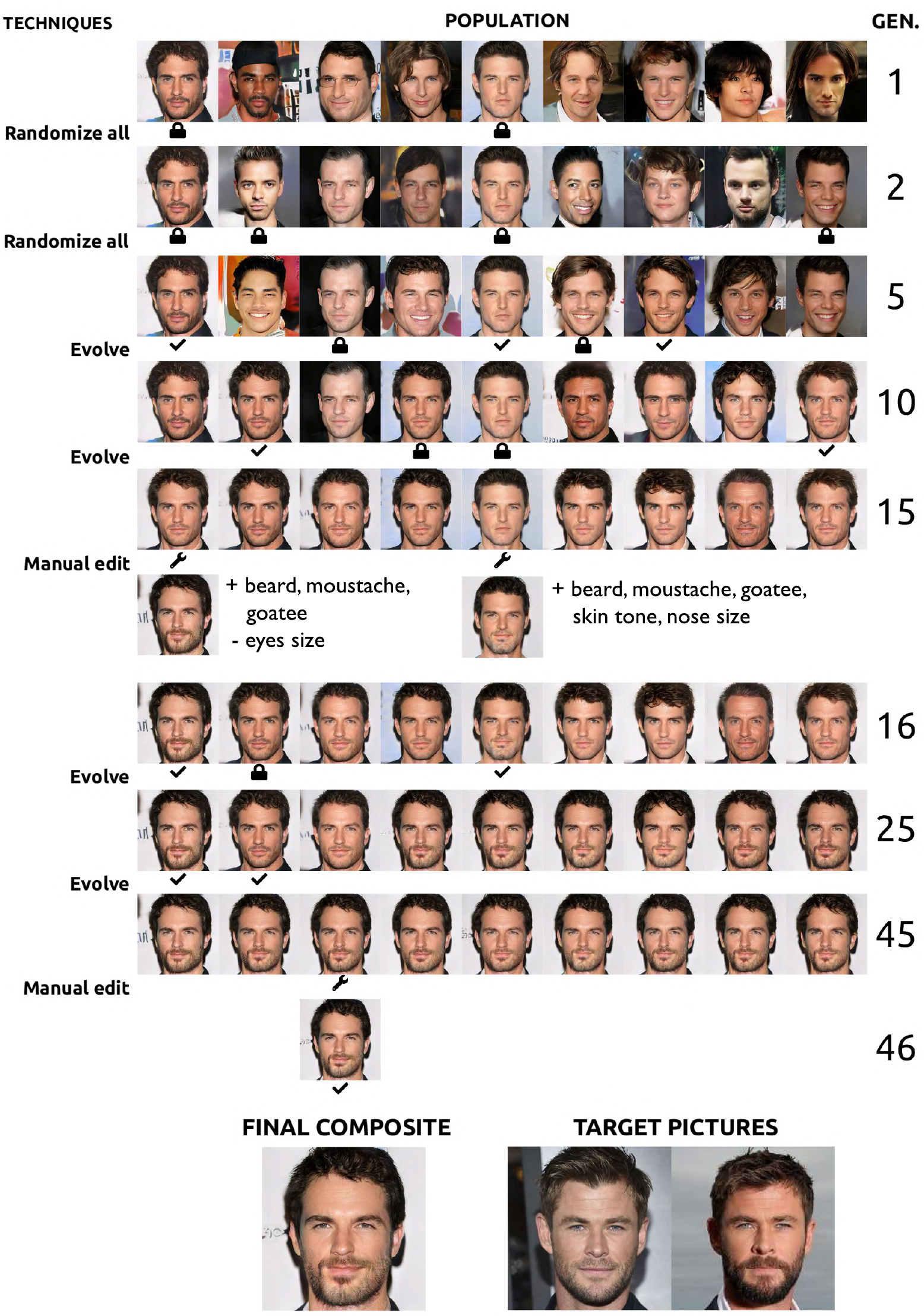}
    \caption[Example session]{Shown is an example of a user session, including the employed user actions  across generations. Given a target (Chris Hemsworth), CG-GAN was used to construct the composite in 46 generations, taking approximately 35 minutes. The user employed randomization, evolution and manual edits. Generations skipped between adjacent rows indicate the same functionality has been used multiple times.}
    \label{fig:workflow}
\end{figure}

\begin{figure}[htb!]
    \centering
    \includegraphics[width=1\linewidth]{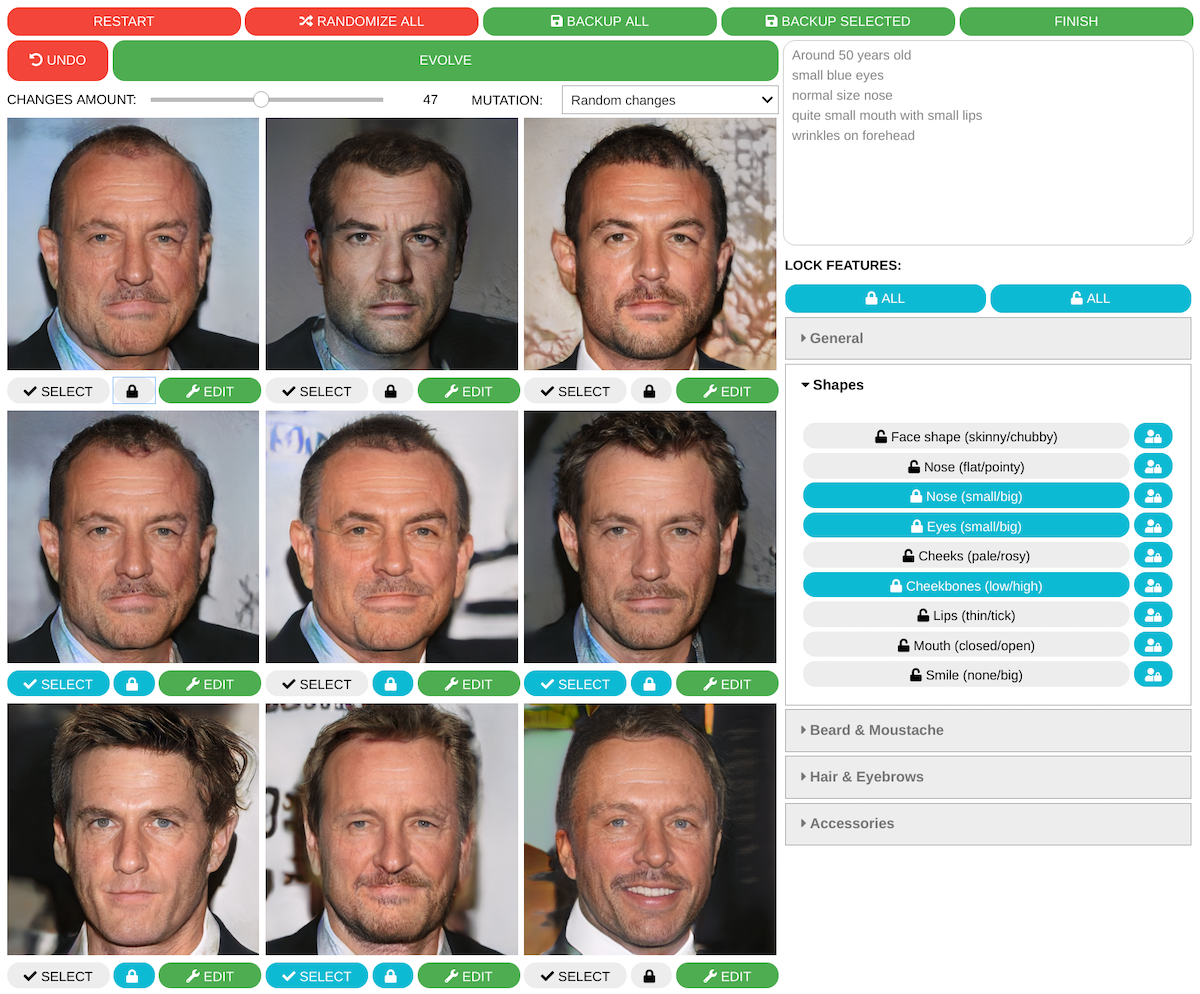}
    \caption{The main UI supports (1) randomization, (2) three kinds of mutation, (3) to lock and unlock features, and (4) to lock, select and manually edit faces. Next to the feature \textit{locks} are also the \textit{smart locks}. Features are changed according to the selected mutation type and a slider controller the amount of change.}
    \label{fig:UI_main_panel}
\end{figure}

A session starts with nine random faces that resemble gender and age chosen in a start-up panel  (Figure~\ref{fig:UI_main_panel}). The user can \textit{select} images to evolve them, \textit{lock} them to preserve them, create new random faces or choose to manually \textit{edit} them. 
Selected images are kept and used as parents for the next generation. Locked images are only kept unvaried. 
The first \textit{free} image (not locked nor selected) is replaced by the crossover-generated offspring, which is the exact average of the selected faces' latent vectors. All the remaining \textit{free} images are replaced with images created by mutating their underlying latent vectors.

Three mutation types are available: \textit{random changes}, \textit{one unlocked feature} and \textit{every unlocked features}. 
\textit{Random changes} adds Gaussian noise to the latent generator input, causing random changes in the faces proportionally to the chosen amount of noise that can be adjusted using a slider. The same applies to all other mutation types.
The \textit{random changes} mutation does not differentiate what aspects of the faces are changed. 
\textit{One unlocked feature} and \textit{every unlocked features} act on specific traits. The user can choose features that should or should not change. 
\textit{One unlocked feature} changes exactly one feature by a certain amount. \textit{Every unlocked feature} changes all the unlocked features by a lower amount, which is proportional to the desired \textit{changes amount} and inversely proportional to the number of features that are being changed. The amount depends on a Gaussian distribution with $\mu$ and $\sigma$ defined as:
\begin{equation}
    \label{math:changes_ampunt}
    \mu = \frac{20 \times \textrm{desired\_changes\_amount}}{F} \qquad \sigma = \frac{\mu}{3}
\end{equation}
\newline where $F$ is set to $1$ in case of \textit{one unlocked feature} and
$
   F =  \min( \max(1, (0.8 \times \#\textrm{unlocked\_features})), 8)
$, 
in case of \textit{every unlocked feature}. The specific values were fine-tuned through prior experimentation. 

As an example, let's assume all features are locked but three: beard, hair length, hair color. \textit{One unlocked feature} will select one of them so that the output face will have either a different beard, hair length or color. \textit{Every unlocked feature} instead will change all those by a lower amount, so that the output has both slightly different beards, hair length, and colors. 

\subsubsection{Advanced editing.}

At any time, the user can manually \textit{edit} a face, by accessing the dedicated panel. The \textit{changes amount} slider and feature locking system work analogously to the ones in the \textit{Main panel}. The additional functionality is the possibility of acting over single feature axes, using a \textit{-} and \textit{+} button. 
A preview is updated at each modification and presets can be saved and loaded. The user can  \textit{save} the changes to overwrite the selected individual in the \textit{Main panel}, which can be further edited or evolved. 

\subsubsection{Exporting results.}

\begin{figure}[htb!]
    \centering
    \includegraphics[width=1\linewidth]{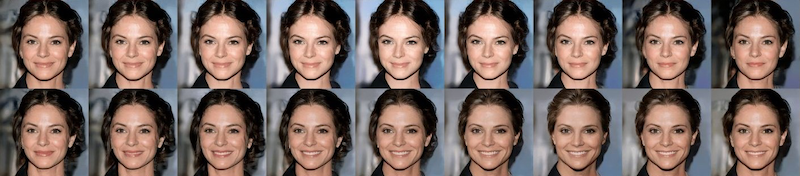}
    \caption[Animated GIF sprite-sheet]{Sequence of frames of an animated composite. 
    }
    \label{fig:animation_spritesheet}
\end{figure}

\begin{figure}[htb!]
    \centering
    \includegraphics[width=1\linewidth]{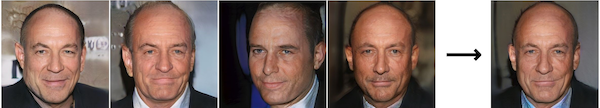}
    \caption[Merged sessions]{A merged session showing four composites (left) and their merged composite (right). The merge composite is generated by taking the average of all latent vectors.}
    \label{fig:merged_sessions}
\end{figure}

When the user is satisfied with the produced results, the session can be concluded and results exported by selecting \textit{finish} after having selected one or more images. In case of multiple selections, an animation that interpolates between  them is created. Figure~\ref{fig:animation_spritesheet} shows a sprite-sheet portion of such an animated GIF. The user can navigate through the frames and export the whole animation or any frame. In the case of our user study we create a single image out of an animation, which is the average of all the selected latent vectors (Figure~\ref{fig:merged_sessions}). 

\subsubsection{Merging multiple witnesses' sessions.}

If more witnesses create composites, it can be useful to present a combined  composite to the public \cite{morph_composites}. CG-GAN allows to load all data of a case with multiple witnesses, and the combined composite is computed based on a simple or weighted average of the underlying latent vectors. If weighted, composites can be assigned weights based on the type of witness\footnote{Witness types are defined as Active, Passive and Inactive according to The Police Composite Sketch \cite{pcs}.} that created them. This feature was not extensively evaluated here but offers an exciting opportunity for future work in composite sketch generation. 

\section{Experiments}
\begin{figure}[htb!]
    \centering
    \includegraphics[width=0.7
    \linewidth]{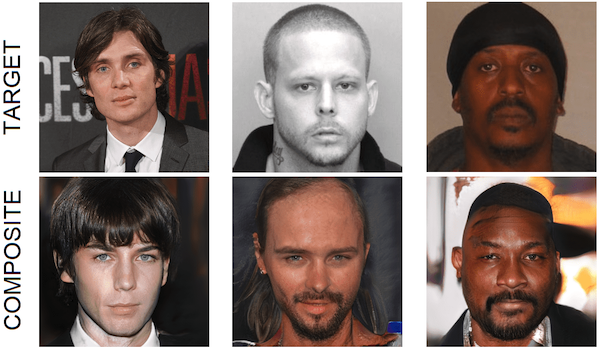}
    \caption[Preliminary testing session results]{Some results from a preliminary testing session. On top are the targets, on the bottom the built composites.}
    \label{fig:preliminary_test_results}
\end{figure}

A preliminary testing phase was performed, which aimed at understanding how non-experts interact with the system. The study involved a diverse group of users and allowed us to fine-tune the interface; some results are shown in Figure~\ref{fig:preliminary_test_results}.

For the final user test, participants were divided in two groups: 13 \textit{constructors} created composites and 55 \textit{evaluators} evaluated them\footnote{Names \textit{constructor} and \textit{evaluator} are inherited from the gold-standard protocol for laboratory evaluation \cite{goldStandard}.}. Participants were recruited among friends, family and students in order to ensure variation in age, habits and skills. 

Regarding the construction of the composites, recall issues were avoided by letting users check the target over the whole duration of the test session.  This choice is driven by our goal of evaluating the potential of a novel approach, rather than evaluating a realistic usage. Indeed a realistic usage scenario would involve factors that could bias the evaluation; it would be harder to pinpoint if a failure of the system is due to the CG-GAN not being capable to create a composite, or to the user having difficulty in remembering the face.

Target images for the user to recreate were generated using the GAN's generator:   Latent vectors were randomized to export four faces not containing graphical imperfections (Figure~\ref{fig:all_results}). Strict rules were set on the timing: every user had 15 minutes to try the software before starting. The sessions lasted up to 30 minutes and data was collected every 10 minutes, to compare different phases of the process.

Testers filled in a survey after every test session. The questions aimed at determining which functions were used, the level of understanding of such functions and the users' satisfaction with their result. The questionnaire, its results, a demo and the code for the experiments in this paper can be found at: \url{https://github.com/LuisaZurlo/CG-GAN}. 

\subsection{Evaluation metrics}
The objective evaluation of the system is a complex task as it depends on subjective recognition abilities \cite{facialCompositeSystemReview}.  To overcome this issue we use two separated evaluation metrics:

\textbf{Similarity score:}
27 \textit{evaluators} assigned a similarity score from 0 to 100 to each composite. We emphasized that the score is meant to represent how likely the images are to depict the target person. 

\begin{figure}[htb!]
    \centering\includegraphics[width=0.7\linewidth]{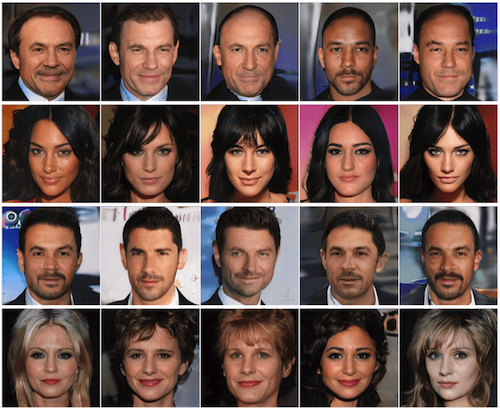}
    \caption[Reasonably different samples]{Different faces shown to users in the \textit{recognition} experiment. Each row represents a line-up, composed of a target and four variations.}
    \label{fig:reasonably_different_samples}
\end{figure}

\textbf{Recognition Rate:}
This measure tests the real purpose of the composites. Each tester (28 total) was given four sets of images. Each set composed of one composite and a lineup of five suspects (Figure ~\ref{fig:reasonably_different_samples}). The users ranked the suspects based on their similarity to the composite. Since the context needs accurate predictions, we only considered the first choice, using the ranking to calculate a recognition rate $r$:
\begin{equation}
  r = \frac{\#\,\textrm{Rank 1} }{\#\,\textrm{Votes}} \times 100
  \label{math:Recognition_Rate}
\end{equation}

It is important to note the potential bias of the metric, as it depends on the shown lineup and the user's subjective perception.
To mitigate this bias, the target images (Figure ~\ref{fig:reasonably_different_samples}) were generated to look \textit{reasonably different} by taking the original GAN-generated image and producing four variations of it by adding a fixed amount of Gaussian noise to the original latent vector. These  reasonably different faces share the majority of common traits, but are  distinguishable from each other. We discarded images that only differ based on light, exposure, pose, or a too limited set of traits.

\section{Results}

\begin{figure}[htb!]
    \centering\includegraphics[width=0.85\linewidth]{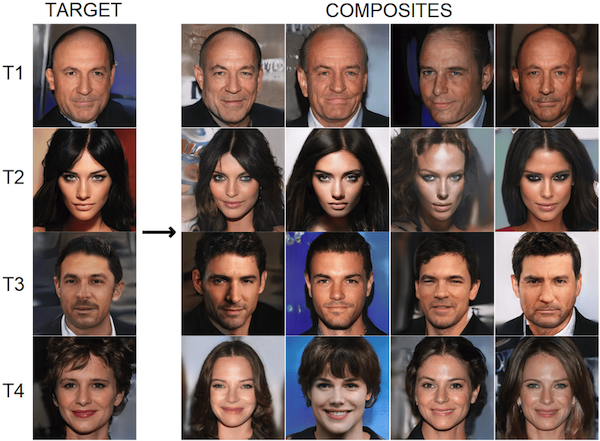}
    \caption[Targets vs. constructed composites]{Target samples (left) and the relative composites built by constructor testers (right).}
    \label{fig:all_results}
\end{figure}

\begin{figure}
    \centering\includegraphics[width=0.77\linewidth]{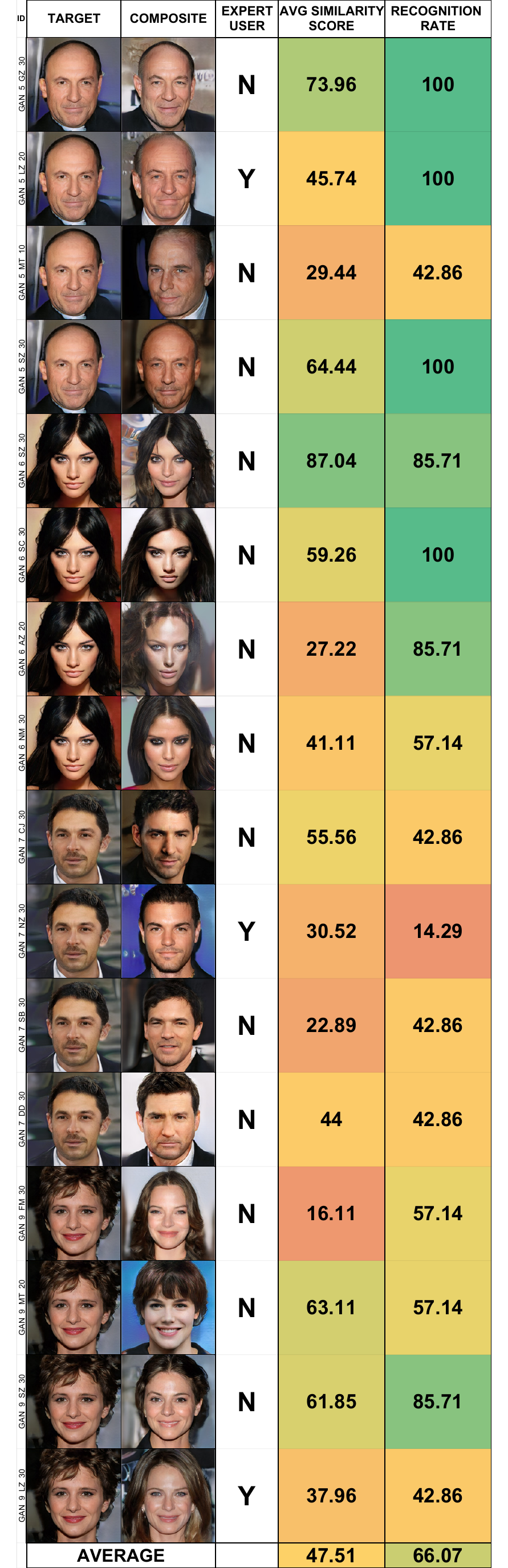}
    \caption[Scores of all the user-generated composites.]{Each row shows the target  image and composite together with an indication if it was created by an expert user. The next column shows the average similarity scores given by the  evaluators, followed by the percentage of users that recognized the particular composite correctly.
    }
    \label{fig:results_all_scores}
\end{figure}

Figure~\ref{fig:all_results} shows the 16 composites generated by the different users for the four selected target samples. Scores are shown in Figure~\ref{fig:results_all_scores}. Overall the results were promising, with 10 out of 16 composites recognized by at least 50\% of all users, and  7 out of 16 composites recognized by at least 75\% of all users.  The average  \textit{similarity} score was 47.51\%.

Comparing the results directly to the previous baseline by \citeauthor{deepInteractiveEvolution}~\shortcite{deepInteractiveEvolution} is difficult since they did not include any recognition test in their experiments. However, users in their study reported a relatively low score in their ability to reproduce a given target face (an average score of 2.2 out of 5). Also, because the users could not freeze any features, they had to find less efficient workaround strategies that might have lead to some frustration. Visually, the quality of the resulting faces is also less convincing (Figure ~\ref{fig:previous}), while the fine-grained control of CG-GAN allowed users to create face composites matching the target image to a  higher degree. Overall, users rated their experience with our system with a score of 7.31 (out of 10) and reported a score of 7.18 when asked how confident they were with the different available functions in the system. Users rated the accuracy of the resulting composite with 6.81.

Overall we  noticed that after some generations  users started to find traits that, according to their perception, resembled the target. Interestingly, the four targets have noticeably different scores. Images resembling target 1 are  in general much more similar to the target compared to the ones representing targets 3 and 4 (Figure~\ref{fig:all_results}). These results were mirrored by the results from the questionnaire, in which users noted that some targets are easier to recreate than others. Very high variance on votes for the same composites proves that the similarity evaluation is very dependent on the particular person and focuses on different facial aspects, a factor well known from police sketching \cite{pcs}. 

\begin{figure}[htb!]
    \centering\includegraphics[width=0.9\linewidth]{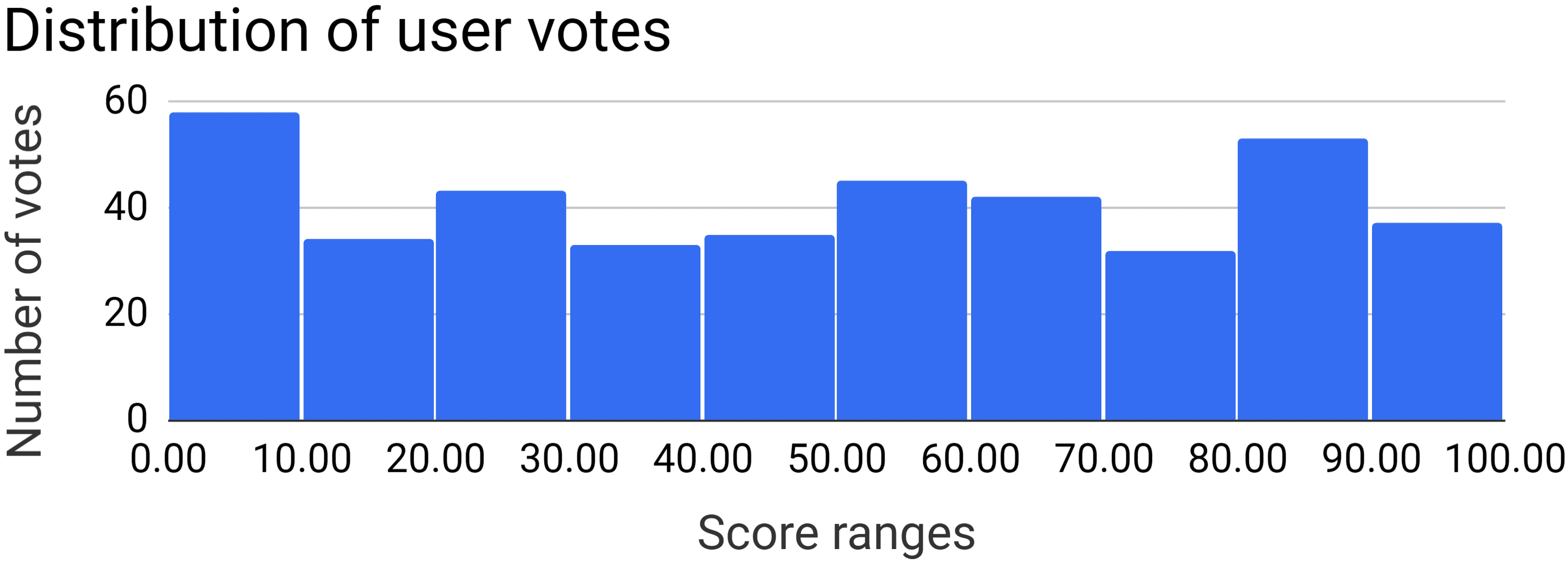}
    \caption[Distribution of user votes]{Histogram showing the distribution of similarity scores given by all users for all 16 result composites. }
    \label{fig:results_user_score_distribution_all_composites}
\end{figure}

Analyzing the overall similarity scores given by the users (Figure \ref{fig:results_user_score_distribution_all_composites}) suggests that voters tended to assign scores that were either low or high instead of following the expected normal distribution; they are almost equally divided with the only exception for very low values (range [0,9]). The surprising trend of user-assigned scores motivated a separate test to prove users' consistency over time. After 72 hours, six \textit{evaluators} repeated the score assignment. Interestingly, most votes were similar but some were completely different (50+ difference), with an average difference of above 20. 

\textit{Constructors} who created the composites assigned higher scores than \textit{evaluators} in 75\% of the cases, and for the remaining 25\% the difference is marginal.
Since the high scores assigned by \textit{constructors} are also assigned by some \textit{evaluators},  those people possibly  perceived some aspects differently than other users. This suggests that low scores may be due to \textit{constructors} perception rather than a deficiency of the system. In other words, they did not produce a \textit{better} composite because they were already satisfied.
Similar cases were found while testing DeepIE, indeed authors often disagreed with the users'  choice of the best results \cite{deepInteractiveEvolution}.

It was expected from expert users -- the authors -- to achieve the best results, mirroring previous experiments in IEC~\cite{accelerating_the_evolution}. Surprisingly, the best composites were generated by novice users, and not all of them even familiar with using complex software (Figure~\ref{fig:results_all_scores}). 

\section{Discussion and Future Work}
The promising results could be improved further by training a GAN with a more extensive and diverse training set. For example, if no celebrities have scars, GANs will probably not become capable of representing scars. The same applies to any unique or not widespread feature. 

\textbf{Feature limitations.} Additionally, obtaining feature axes via supervised learning requires significant  training data. The lack of labeled face images prevented possible improvements and additional axes to be learned.
Some workarounds have been attempted to bypass the problem: manual feature research and overlays.
In the first case, we tried to infer axes via trial-and-error, with no remarkable results.  The second approach involved overlays for altering colors over specific areas, e.g.\ colored ellipses over the eyes. Their downside is the missing relation with the genotype, so the modifications could not be translated into a latent vector change and re-evolved.

\textbf{Environmental factors.}
Forensic artists take into consideration conditions that may affect the perception such as lighting, location and time \cite{pcs}. It may be useful to train GANs on specific environment-dependent datasets, to further improve our results.

\textbf{Feature axes improvements.}
The CelebA \cite{celeba} dataset served the purpose of learning 40 features. However, some were marginally relevant and additional ones would be helpful.
Some useful additions could be dimensionalities of the face (width, length, distance between eyes) or its shape (oval, round, square). An interesting improvement would be to learn feature axes from computed measures rather than a labeled dataset, e.g.\ through the usage of facial recognition tools. Measures may include eyes color, emotions, or pose.

\textbf{Perceptual impact of external features}
For a similar application, it was demonstrated that blurring external features such as hair with Gaussian filters allows maintaining context while concentrating on more important inner features \cite{Changing_the_face_of_criminal_identification}. Implementing a similar technique in the CG-GAN system would probably improve its naming rate as for other systems.

\section{Conclusion}

The presented composite creation approach suggests  the promising involvement of generative techniques in this area.  Current composite creation systems mostly belong to two main categories. Some rely on datasets of drawn features that are arranged, moved and stretched to build a face \cite{faces},  which are not holistic and only work with pre-defined features. Others evolve parameters to be used by mathematical functions to create a face  \cite{facialCompositeSystemReview,evofit_paper,efitV} but can only create traits that these functions are designed for through expert knowledge.  CG-GAN, on the other hand, learns to generate whole faces rather than assembling them from sets of components or mathematical rules. Expert knowledge is not necessary and high-quality colored animated composites are produced by directly learning from pictures.  Our results suggest that even inexpert users can produce good composites in a limited time. Most issues appear promisingly simple to solve, granting a likely  accuracy improvement. The applied pre-trained generator network is used as is, however it could be re-trained on a more diverse database or different domains. 

\bibliographystyle{aaai}
\bibliography{bib}

\end{document}